\documentclass[letterpaper, 10 pt, conference]{ieeeconf}  

\IEEEoverridecommandlockouts                              

\usepackage{graphics} 
\usepackage{amsmath} 
\usepackage{amssymb}  
\usepackage{todonotes}
\usepackage{physics} 
\usepackage{algorithm}
\usepackage{xcolor}

\usepackage{wrapfig}
\usepackage{algpseudocode}
\usepackage{subfig}
\usepackage{caption}
\usepackage{cite}
\title{\LARGE \bf
Self-Supervised Learning of Object Segmentation\\ from Unlabeled RGB-D Videos
}

\author{Shiyang Lu$^{1}$, Yunfu Deng$^{2}$, Abdeslam Boularias$^{1}$, Kostas Bekris$^{1}$
\thanks{$^{1}$The authors are affiliated with the Department of Computer Science at Rutgers University, New Brunswick, NJ, 08901, USA. Email: \{shiyang.lu, abdeslam.boularias, kostas.bekris\}@rutgers.edu. $^{2}$This author is affiliated with the Department of Electrical and Computer Engineering at Rutgers University, New Brunswick, NJ, 08901. This work is supported by NSF awards 1734492, 1846043 and 2132972.}
}

\begin{document}

\maketitle


\begin{abstract}
This work proposes a self-supervised learning system for segmenting rigid objects in RGB images. The proposed pipeline is trained on unlabeled RGB-D videos of static objects, which can be captured with a camera carried by a mobile robot. A key feature of the self-supervised training process is a graph-matching algorithm that operates on the over-segmentation output of the point cloud that is reconstructed from each video. The graph matching, along with point cloud registration, is able to find reoccurring object patterns across videos and combine them into 3D object pseudo labels, even under occlusions or different viewing angles. Projected 2D object masks from 3D pseudo labels are used to train a pixel-wise feature extractor through contrastive learning. During online inference, a clustering method uses the learned features to cluster foreground pixels into object segments. Experiments highlight the method's effectiveness on both real and synthetic video datasets, which include cluttered scenes of tabletop objects. The proposed method outperforms existing unsupervised methods for object segmentation by a large margin.
\end{abstract}

\section{Introduction}
Autonomous robots should be able to reason about objects even when human supervision is not available~\cite{lu2022online}. This work considers a setup where a robot navigates in a static environment and passively collects RGB-D videos. The environment contains unknown rigid objects, which rest stably, potentially in cluttered configurations, on supporting surfaces like tabletops. The same object instances can reappear in various scenes with arbitrary 6D poses while being partially occluded by different surrounding objects. This raises the following question: Can a robot learn to correctly segment objects in RGB images given its prior experience of collecting unlabeled videos that contain the involved objects without any human supervision? 
\begin{figure}[t]
    \centering
    \includegraphics[width=\columnwidth]{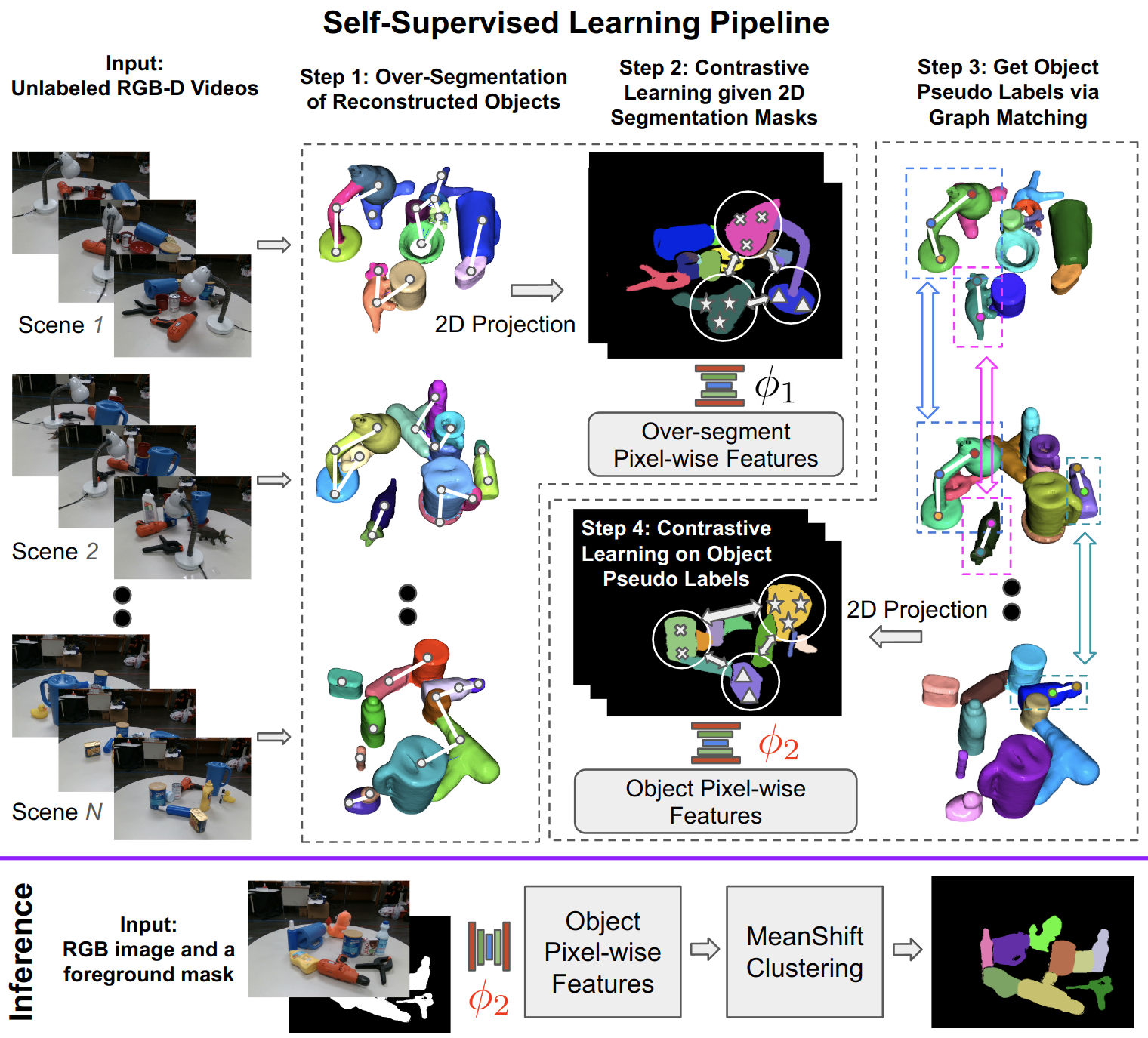}
    \caption{\textbf{Self-Supervised Learning Pipeline.} Top: Step 1. Object over-segmentation on the 3D  reconstruction of each video. Step 2. Generating a distinctive feature for each 3D segment by training an initial feature extractor $\phi_1$ using contrastive learning. Step 3. A graph-matching algorithm operates over the 3D segments and their $\phi_1$ features to identify sets of segments to be grouped as objects. Step 4. Using contrastive learning to train another feature extractor $\phi_2$, which uses as positive examples pixels that belong to the same hypothesized objects. Bottom: Upon inference, given an RGB image and a foreground mask, the network $\phi_2$ generates pixel-wise features as input to a clustering algorithm for object segmentation.}
    \label{fig:pipeline}
    \vspace{-.3in}
\end{figure}

While some objects, such as a coke can or a sugar box, can be relatively easily singled out from a scene due to their characteristic texture or simple geometries, other more complex and articulated objects pose a more serious challenge in the absence of prior knowledge. For example, a lamp can have a concave lampshade, a gooseneck, and a base. It's hard for a robot to tell from a single snapshot if they are individual objects or parts of the same object. This paper argues that robots can address this problem without prior knowledge or human supervision by simply collecting videos of scenes, which can be extended to a lifelong learning process. The idea is to identify object parts that appear simultaneously in the same spatial arrangement in different scenes. The key contribution of this work towards this idea is a graph-matching algorithm that identifies reoccurring sets of neighboring 3D object parts. The overall approach utilizes the results of the matching algorithm to train a pixel-wise feature extractor through contrastive learning, which can be directly consumed by a clustering method for object segmentation.


The proposed pipeline is shown in Fig.~\ref{fig:pipeline}. The learning system receives as input a collection of unlabeled RGB-D videos from multiple scenes and returns a pixel-wise feature extractor $\phi_2$ that can generate distinctive object features for segmentation. The system first performs (over-)segmentation of the 3D point cloud for each input video resulting in 3D segments corresponding to simple geometries. This is done by reconstructing each 3D scene, removing the background using plane detection, and segmenting the scene using existing non-learning, geometry-based methods~\cite{lccp, symseg}. The 3D segments are projected to the 2D frames, and a pixel-wise feature extractor $\phi_1$ is learned to generate distinctive features for each 3D segment through contrastive learning, similar to MaskContrast~\cite{vangansbeke2020unsupervised}. The 3D segments are then abstracted as nodes of a graph, where edges connect adjacent nodes, and nodes are represented by their features according to $\phi_1$. Reoccurring graph patterns are searched across different scenes by a feature-based, sub-graph matching algorithm~\cite{graph_matching}. Graph matching is used here as a mechanism to identify candidate pairs of object instances with multiple parts. The corresponding point clouds of the candidates are registered and matched nodes with low registration scores are pruned out. The remaining matched nodes and the segments represented by those nodes are considered parts of the same object. Then, a second network $\phi_2$ is learned using contrastive learning to generate pixel-wise features of object instances. During inference, given an RGB image and a foreground mask, $\phi_2$ returns pixel-wise object features, which are then clustered  using a variant of the mean-shift algorithm~\cite{comaniciu2002mean, banerjee2005clustering} to output the object segmentation of the input RGB image. 

The experimental results of Section~\ref{experiments} show that the proposed technique significantly outperforms multiple baselines both on real-world and photo-realistic synthetic datasets.

\section{Related Work}\label{related_work}

\textbf{Self-Supervised Learning.}
Many recent studies on self-supervised learning~\cite{simclr, mocov2, simsiam, byol, ICRA2021Juntao} focus on generating a pre-trained model from a large dataset using pseudo labels without human supervision. Given the pre-trained model, downstream tasks (e.g., object segmentation) can be trained with less annotated data. Many of the existing methods require object region proposals that contain one or very few salient objects. This object localization process is usually performed by 2D region proposals~\cite{9196567} and saliency detection~\cite{vangansbeke2020unsupervised} for static images, or object tracking~\cite{wang2015unsupervised, wang2017transitive} and optical flow~\cite{pathak2017learning, liu2021emergence} for videos. Nevertheless, they are not as suitable for cluttered scenes~\cite{purushwalkam2020demystifying} as the errors introduced during region proposal are often too large. Meanwhile, generating a pre-trained model is not ideal as it still requires manually labeled ground truth to finetune for downstream tasks, such as segmentation. This work, on the other hand, focuses on finding object pseudo labels that can be directly used for training.

\textbf{Segmentation without Manual Annotation}. Many unsupervised methods have been proposed for object/scene segmentation. Notable non-learning methods are Mean Shift~\cite{comaniciu2002mean} for image segmentation, and LCCP~\cite{lccp}, SymSeg~\cite{symseg},  CPC~\cite{cpc} for point cloud segmentation. Some efforts have studied RGB-D segmentation for unknown objects, including UOC~\cite{uoc}, UOIS~\cite{uois}, RICE~\cite{rice}, SD-MaskRCNN~\cite{sd-maskrcnn}. These methods typically train a neural network on synthetic data, where ground-truth labels are automatically generated. Fine-tuning these models on real data is nontrivial without manual annotation. Recently, some work has been done on unsupervised semantic segmentation. PiCIE~\cite{Cho_2021_CVPR} uses invariance to photometric effects and equivariance to geometric transformations as an inductive bias. STEGO~\cite{stego} proposes a novel contrastive loss function that encourages features to form compact clusters while preserving their relationships across the corpora.

\textbf{Unsupervised Object Discovery} This work is also related to unsupervised object discovery in general. Some representative works include~\cite{cho2015unsupervised}, which is an object detection method that uses part-based matching with bottom-up region proposals, and ~\cite{vo2021large} that formulate the problem as ranking amenable to distributed methods available for eigenvalue problems and link analysis.

\section{Problem Setup and Notation}\label{problem_setup}
This work addresses the problem of object segmentation given an RGB-D image $I$ of a novel scene, which contains various, potentially cluttered objects. Each object is assumed to have appeared before in a collection of $m$ unlabeled raw RGB-D videos $V = \{V_1, V_2, ..., V_m\}$ that were passively recorded by a mobile robot while observing static scenes. There are $K$ unique object instances $O = \{O_1, O_2, ..., O_K\}$ that the robot can observe. Image frame at time $t$ in the training $i$-th video $V_i$ is denoted as $I^t_i$, i.e., $V_i = (I^1_i , I^2_i, ..., I^{N_i}_i)$, where $N_i$ is the total number of frames in $V_i$. Each frame $I^t_i$ is composed of a color and a depth image denoted as $I^t_i.color$ and $I^t_i.depth$ respectively. Each frame $I$ of a video, as well as the test image, may contain only a subset of the object set $O$, denoted as $O^I \subseteq O$. Background objects, such as furniture, are assumed to be known a priori and are automatically removed from the RGB-D images. No additional information, such as the number of objects in each scene, or labels that help with segmentation is available.

\section{Object Segmentation during Testing}
\begin{figure}[h]
    \vspace{-.125in}
    \centering
    \includegraphics[width=\columnwidth]{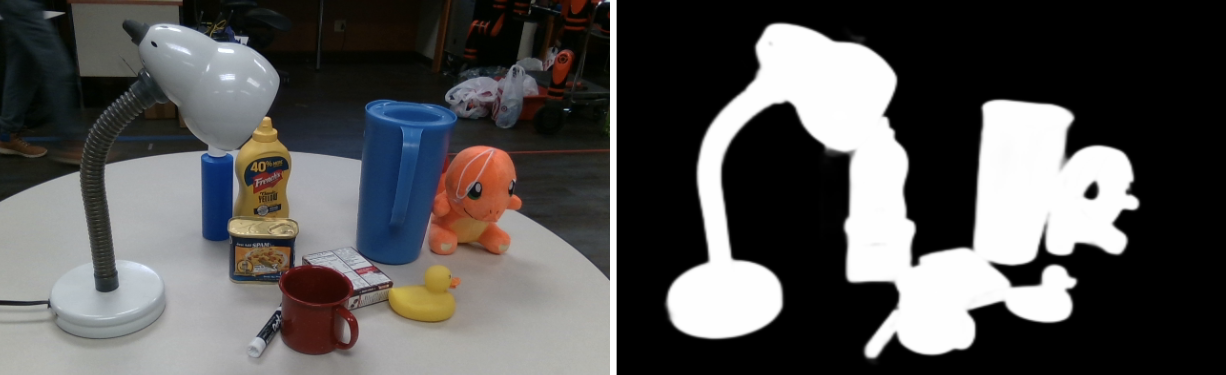}
    \vspace{-0.2in}
    \caption{Example of saliency detection~\cite{basnet} for generating the foreground mask (right) given RGB input (left).} 
    \label{fig:saliency}
    \vspace{-.125in}
\end{figure}

During online inference, a learned function $\phi_2(x): \mathcal{X} \rightarrow \mathcal{Z}$ is used to extract a normalized feature vector $z$ for each pixel $x$ in the given test RGB image. A clustering algorithm is applied over the foreground pixels based on their extracted features. This foreground mask can be generated with a saliency detection method, such as BasNet~\cite{basnet} and DeepUSPS~\cite{deepusps}. See the example of Fig.~\ref{fig:saliency}.

Since the number of objects in a given image is not known, clustering methods, such as K-Means, which require this input, are not applicable. Thus, this work adapts a variant of the von Mises-Fisher mean-shift~\cite{banerjee2005clustering} for feature clustering. This variant not only considers the feature similarity of pixels but also the 2D proximity of pixels. A simple post-processing step is applied so that pixels in a feature cluster are separated if the corresponding pixels do not form a connected component. The online inference is computationally efficient ($\sim$ 6fps). The detailed training process of $\phi_2$ is described in the following section.

\section{Self-Supervised Learning Pipeline}\label{pipeline}

\subsection{Video Pre-processing}
Given a set of unlabeled RGB-D videos of static objects, a dense reconstruction method, where KinectFusion~\cite{kinectfusion} is adopted in this work, is used to estimate the camera pose for each frame and reconstruct a 3D scene for each video. The foreground 3D region of interest is extracted by removing known background objects, e.g. table. And the 2D foreground masks during training are generated by projecting the filtered point cloud to individual 2D frames. 3D reconstruction is employed here instead of individual depth images for multiple reasons: a) It allows to generate pseudo labels on the reconstructed point cloud once and then acquiring consistent labels for segments on all of the frames of each video via 3D to 2D projection, b) It increases the visibility of each object given multiple views, which reduces issues due to occlusions that can arise for individual viewpoints, c) It avoids having to find correspondences between segments across frames, which can be error-prone. 

\subsection{Object Over-Segmentation}
Each reconstructed 3D scene is over-segmented into a set of small segments corresponding to simple geometries. Ideally, this over-segmentation should be consistent across different scenes, where consistency means that the same object surface is segmented in the same way across different scenes. This work first adopts LCCP~\cite{lccp} for this purpose. LCCP is a non-learning segmentation method based on the local convexity of point clouds. It is able to segment convex objects, such as a box, into a whole, and segment complex objects into small, approximately convex parts. During the development of this work, it is found that LCCP can fail on certain objects with a large concave surface, such as a bowl.

\begin{figure}[h]
    \centering
    \includegraphics[width=0.48\textwidth]{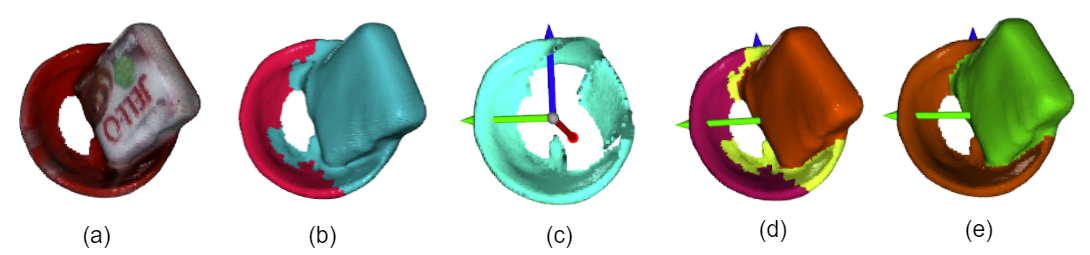}
    \caption{\textbf{Splitting and Merging Segmentation Results of LCCP and SymSeg.} (a) Reconstructed point cloud of a ``Jell-O'' box in a bowl. (b) Irregular segments generated by LCCP due to notable concavity. (c) Detected rotational axis and segments generated by SymSeg. (d) Splitting segments. (e) Merging segments.}
    \label{fig:split_and_merge}
    \vspace{-.3in}
\end{figure}

To alleviate this issue, SymSeg~\cite{symseg} is introduced to assist with improving (over-)segmentation quality for symmetric concave objects. SymSeg first detects symmetries in a point cloud and then uses them for segmentation. If no symmetry is detected, then SymSeg is not applied. In this work, only rotational symmetries are used, since reflection symmetries often lead to under-segmentation in clutter. A simple strategy to fuse the results of LCCP and SymSeg is illustrated in Fig.~\ref{fig:split_and_merge}. A 3D segment generated by LCCP is first split into multiple parts if an overlapping symmetric segment is detected by SymSeg. LCCP segments are then merged if they appear in the same symmetric segment of SymSeg, with an exception when both parts share the same symmetric axis (e.g. a bottle standing in a bowl). Small segments with less than 200 points are pruned. Note that this work does not argue that this combination is the best for over-segmentation, but they work sufficiently well as a submodule in the pipeline. Any recent progress in class-agnostic part segmentation may benefit this work.

\subsection{Contrastive Learning}\label{contrastive_learning}
To generate a distinctive feature for each 3D segment, contrastive learning is adopted to train a pixel-wise feature extractor $\phi$. The features will be first used during graph matching and later in object segmentation. 3D segments are projected to 2D pixels using estimated camera poses of video frames. Given the assumption that points in the same segment belong to the same object and their features should be similar, the embeddings of a pair of pixels $(i, j)$ are pulled closer if they belong to the same 3D segment, or pushed away otherwise. In practice, $P \gg N$, where $P$ is the number of foreground pixels per image, and $N$ is the number of segments per scene. The number of pixel pairs is $O(P^2)$, which results in a computationally expensive sampling process. To alleviate this issue, the strategy from MaskContrast~\cite{vangansbeke2020unsupervised} is adopted to reduce the computational complexity to $O(PN)$. Let $M_i$ be the set of pixels from segment $i$. Define the mean pixel embedding of $M_i$ as $\bar{z}_i = \frac{1}{|M_i|}\sum_{k =1}^{|M_i|} z^k_i$. The mean feature $\bar{z}_i$ is used to represent the features of all pixels in $M_i$. Given positive pairs $(z_k, \bar{z}_i)$, for $k \in M_i$, and negative pairs ($z_k, \bar{z}_j)$, for $ k \not\in M_j$, the contrastive loss for each foreground pixel $k$ is defined as:
$\mathcal{L}_{k} = -log\frac{exp(z_k \cdot \bar{z}_i) / \tau}{\sum_{j=1}^N exp(z_k \cdot \bar{z}_j / \tau)}$,
where $\tau$ is a temperature hyperparameter that has a constant value of $0.07$ in all our experiments (as in MoCo~\cite{mocov2}). The mean feature $\bar{z}_i$ is used to represent the 3D segment $i$ after training. 

\begin{figure*}[h]
    \centering
    \includegraphics[width=0.8\textwidth]{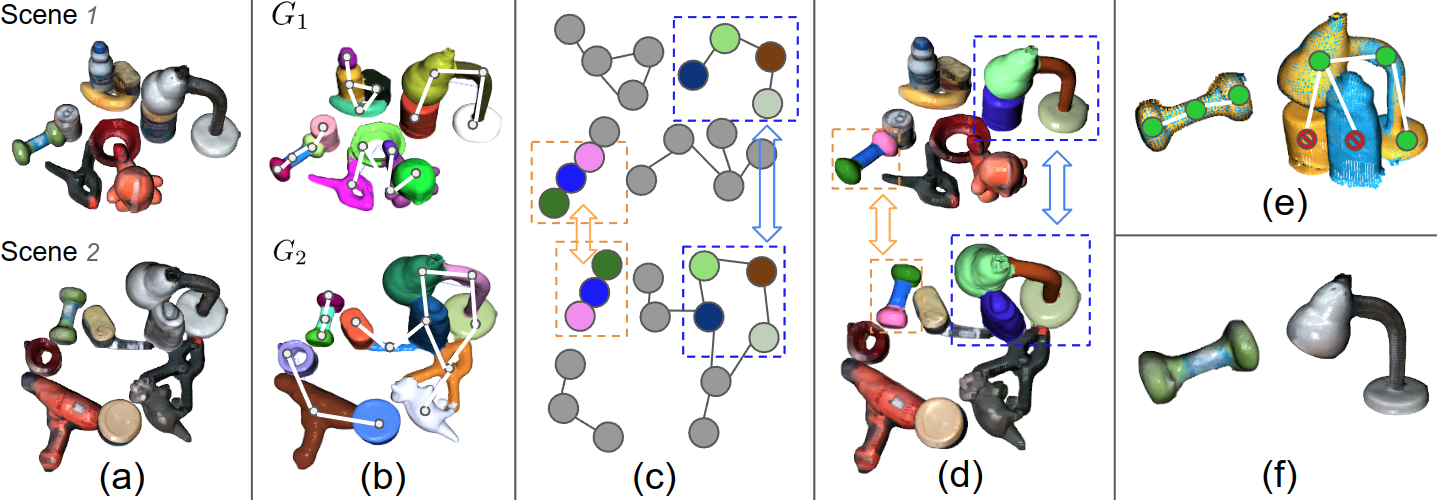}
    \caption{\textbf{Graph Matching.} (a) Colored 3D reconstruction of two scenes. (b) 3D segments, and corresponding graphs, resulting from over-segmentation of the point clouds. (c) Matched nodes from graph matching are highlighted. Parts of the ``dumbbell" and the ``lamp" are correctly matched, while the ``bleach cleanser" and the ``master chef can" are incorrectly matched. (d) Matched connected components are shown in the same color. (e) 3D point cloud registration of matched components. (f)  Segments/nodes that do not register well (such as the ``bleach cleanser" and the ``master chef can") are rejected before merging the remaining segments into objects.} 
    \label{fig:graph_matching}
    \vspace{-.25in}
\end{figure*}

\subsection{Graph Matching}\label{graph_matching}
For each scene, a graph is generated by linking each 3D segment to a node and defining an edge between every two segments that share a boundary in the 3D space. Reoccurring subgraphs across multiple graphs (i.e., scenes) are hypothesized objects since it is unlikely for two separate objects to always appear in the same poses relative to each other in all the scenes encountered by the robot. To find these reoccurring subgraphs, this work adapts an algorithm based on error-tolerant, minimum-cost subgraph matching~\cite{graph_matching}. Given a source graph $G_1=(\mathcal{V}_1, \mathcal{E}_1)$ and a target graph $G_2=(\mathcal{V}_2, \mathcal{E}_2)$, the algorithm identifies a subgraph $S \subseteq G_2$, that minimizes the cost of matching $S$ to a subgraph in $G_1$ given both structural and feature distortions. This optimization problem is formulated into a binary linear program (BLP) with the following objective function:
\begin{equation}
    \vspace{-.1in}
    \footnotesize
    \begin{split}
        J = \min_{x, y, \alpha, \beta} \Big( & \sum_{i \in \mathcal{V}_1}\sum_{k \in \mathcal{V}_2} x_{i, k} \cdot c(i \rightarrow k)) + \sum_{i \in \mathcal{V}_1} \alpha_i \cdot c(i \rightarrow \epsilon) + \\
        & \sum_{ij \in \mathcal{E}_1}\sum_{kl \in \mathcal{E}_2}y_{ij, kl} \cdot c(ij \rightarrow kl) + \sum_{ij \in \mathcal{E}_1} \beta_{ij} \cdot c(ij \rightarrow \epsilon)\Big) \label{eq:j1}
    \end{split}
    \vspace{-.1in}
\end{equation}
where $x_{i,k}$, $y_{ij,kl}$, $\alpha_i$ and $\beta_{ij}$ are binary variables to optimize: $x_{i,k}$ expresses the mapping of node $i$ from source graph $G_1$ to node $k$ from target graph $G_2$; $y_{ij,kl}$ expresses the mapping of edge $(i, j)$ from source graph $G_1$ to edge $(k,l)$ from target graph $G_2$; deletion variables $\alpha_{i}$ and $\beta_{ij}$ are set to be 1, if node $i$ and edge $(i,j)$ are deleted, respectively, according to the match. This can happen when the 3D segment corresponding to node $i$ in the source scene does not appear anywhere in the target scene. The cost of node mapping $c(i \rightarrow k)$ is defined as the cosine distance between the corresponding segment features $\bar{z}_i$ and $\bar{z}_k$. The cost of edge mapping $c(ij \rightarrow kl)$ is set to 0 here as no edge feature is provided. The cost $c(i \rightarrow \epsilon)$ (or $c(ij \rightarrow \epsilon)$) expresses the cost of not assigning a vertex (or edge) of the source graph to a vertex (or edge) in the target graph. Both costs are set to 0.1 in the implementation, which can be viewed as thresholds for rejection.

\begin{subequations}
\footnotesize
\begin{equation}
    \sum_{k \in \mathcal{V}_2}x_{i, k} \leq 1, \forall i \in \mathcal{V}_1 ;\hspace{.1in} \sum_{i \in \mathcal{V}_1}x_{i, k} \leq 1,\forall k \in \mathcal{V}_2 \label{eq:a}\\ 
\end{equation}
\vspace{-.1in}
\begin{equation}
    \alpha_i = 1 - \sum_{k \in \mathcal{V}_2}x_{i,k}, \forall i \in \mathcal{V}_1;\hspace{.1in} \beta_{ij} = 1 - \sum_{kl \in \mathcal{V}_2}y_{ij, kl}, \forall ij \in \mathcal{E}_1 \label{eq:b}\\
\end{equation}
\vspace{-.1in}
\begin{equation}
    \begin{split}
    \sum_{l \in \mathcal{V}_2\textrm{ s.t. } kl\in\mathcal{E}_2} y_{ij, kl} \leq x_{i,k}, \forall ij \in \mathcal{E}_1, \forall k \in \mathcal{V}_2;\hspace{.1in} \\ 
    \sum_{k \in \mathcal{V}_2 \textrm{ s.t. } kl\in\mathcal{E}_2} y_{ij, kl} \leq x_{j,l}, \forall ij \in \mathcal{E}_1, \forall l \in \mathcal{V}_2 \label{eq:c}
    \end{split}
\end{equation}
\vspace{-.1in}
\begin{equation}
    \begin{split}
    x_{i,k} \in \{0,1\}, \forall i \in \mathcal{V}_1, \forall k \in \mathcal{V}_2;\hspace{.1in}\\
    y_{ij,kl} \in \{0,1\}, \forall ij \in \mathcal{E}_1, \forall kl \in \mathcal{E}_2 \label{eq:d}
    \end{split}
\end{equation}
\end{subequations}

The objective function is subject to constraints \ref{eq:a}-\ref{eq:d}. Constraints \ref{eq:a} indicate that each node from either graph can be mapped to at most one node from the other graph. Constraint \ref{eq:b} indicates that if a node in $G_1$ is not matched to any vertex in $G_2$, then it must be deleted. Constraints \ref{eq:c} indicate that an edge should be mapped only if the corresponding vertices are also mapped. The constraints \ref{eq:d} ensure that the variables to be optimized are binary variables. Note that constraints \ref{eq:b} are implicitly respected given constraints \ref{eq:a} and \ref{eq:c}. Thus, the deletion variables $\alpha_i$ and $\beta_{ij}$ in equation \ref{eq:j1} can be removed to reduce the search space resulting in the following objective function: \vspace{-.05in}
\begin{equation}\label{objective_function}
    \footnotesize
    \begin{split}
        J = \min_{x, y} \Big(& \sum_{i \in \mathcal{V}_1}\sum_{k \in \mathcal{V}_2} x_{i, k}\big(c(i \rightarrow k)) - c(i \rightarrow \epsilon)\big)\\
        + & \sum_{ij \in \mathcal{E}_1}\sum_{kl \in \mathcal{E}_2}y_{ij, kl}\big(c(ij \rightarrow kl) - c(ij \rightarrow \epsilon)\big)\\
        + & \sum_{i \in \mathcal{V}_1}c(i \rightarrow \epsilon) + \sum_{ij \in \mathcal{E}_1}c(ij \rightarrow \epsilon)\Big)\\
    \end{split}
\end{equation}
Furthermore, since the 3D segment graphs are undirected, constraints \ref{eq:c} can be further combined into a single one, i.e.
\begin{equation}\label{eq:c_rewrite}
    \vspace{-.05in}
    \footnotesize
    \sum_{l \in \mathcal{V}_2 \textrm{ s.t. } kl\in\mathcal{E}_2} y_{ij, kl} \leq x_{i,k} + x_{j,k}, \hspace{0.1in}\forall ij \in \mathcal{E}_1, \forall k \in \mathcal{V}_2
\end{equation}
The final formulation is to minimize the objective function according to Eq.~\ref{objective_function} given the constraints of Eqs.~\ref{eq:a}, \ref{eq:c_rewrite} and \ref{eq:d}. Each weakly connected component in every 3D scene graph is matched against all the weakly connected components in all other scenes. Note that source and target graphs are not commutative, i.e., $J(G_1, G_2) \neq J(G_2, G_1)$.

\subsection{False Matches Pruning}\label{False Matches Pruning}
This optimization  returns matched node pairs with minimum cost, which are used to find matched subgraphs. Three criteria are used to prune falsely matched subgraphs.\\ {\bf 1.} A threshold $T_1$ is set so that only pairs of subgraphs with a total matching cost $J$ below this threshold are considered valid matches. The implementation sets $T_1$ to be proportional to $N = |\mathcal{V}_1| + |\mathcal{E}_1|$, i.e., $T_1 = \alpha N$, because the total cost is a sum over both node and edge costs. A large graph is likely to generate a higher cost than a small one. The empirical ratio $\alpha$ is set to be $0.1$.\\
{\bf 2.} The matching pairs of subgraphs that satisfy criteria 1 are filtered by RANSAC-based point cloud registration~\cite{ransac}. An empirical threshold $T_2 = 0.9$ is set for both precision and recall of the registered point cloud of each pair of matched nodes. If a pair of nodes does not meet this threshold, the nodes will not be merged with other nodes in their respective subgraphs, as shown in Fig.~\ref{fig:graph_matching}.\\
{\bf 3.} If an object is found to be isolated in a scene, which is performed automatically, then other segments should not be merged with this object. To achieve this, if a source graph $G_1 = (\mathcal{V}_1, \mathcal{E}_1)$ is matched with a subgraph $S$ of the target graph $G_2$, such that $S = (\mathcal{V}', \mathcal{E}'), |\mathcal{V}_1| = |\mathcal{V}'|$, then any matched subgraph $\hat{S} = (\hat{\mathcal{V}}, \hat{\mathcal{E}}) \subseteq G_2$ that contains all the nodes of $S$ as a proper subset, i.e., $\mathcal{V}' \subset \hat{\mathcal{V}}$, will be filtered.

Once the pruning is completed, the remaining nodes in the matched graphs are merged. The merged segments are considered objects and projected back to each frame of the video as object masks for training. The training process for the network $\phi_2$ is the same as in Sec.~\ref{contrastive_learning}.

\section{Experimental Results}\label{experiments}

\subsection{Dataset}
\begin{figure}[h]
    \centering
    \includegraphics[width=0.48\textwidth]{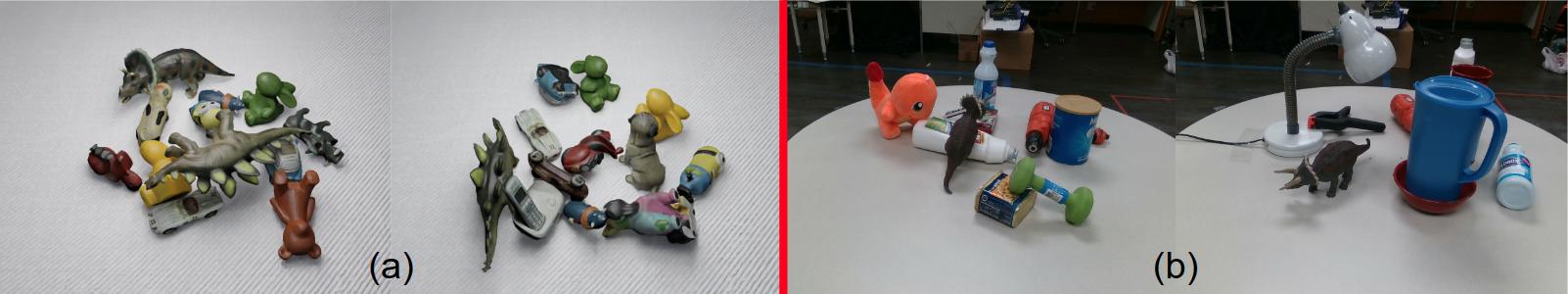}
    \vspace{-0.2in}
    \caption{\textbf{Datasets.} (a) A photo-realistic synthetic dataset of 36 videos using all toys from the HomebrewedDB dataset~\cite{homebrewedDB} rendered by BlenderProc~\cite{blenderproc}. (b) A manually collected real dataset with 30 videos and 20 unique objects.}
    \label{fig:dataset}
    \vspace{-.2in}
\end{figure}

Two video datasets are used for experiments and evaluation, which mimic the behavior of a robot navigating around a table with cluttered objects on top of it~\cite{geofusion}. Some sample frames are shown in Fig.~\ref{fig:dataset}. The first one is a physics-aware photo-realistic video dataset generated by BlenderProc~\cite{blenderproc}. All 17 toys from the HomeBrewedDB dataset~\cite{homebrewedDB} are selected to create random scenes in simulation. For each scene, 12 unique toys are randomly selected and dropped on a surface. A virtual camera captures a video of each scene by rotating around it at a fluctuating height. In total, 36 videos are collected with 800 frames each. The second dataset is collected in real environments with 20 unique objects. Among them, 13 objects are from the YCB dataset~\cite{ycb_dataset}, and 7 objects with more complicated geometries: ``lamp", ``dumbbell", ``toy dinosaur", ``toy duck", ``Clorox bottle", ``detergent", and ``toy Charmander" are included. Each of these non-convex objects contains several sub-parts of varying shapes. The real dataset contains 30 videos with 10 unique objects and $\sim$1000 frames each. The ground-truth segmentation for the real dataset is manually annotated in 3D space for each scene and then projected in 2D frames. Background removal in these scenes is performed using RANSAC~\cite{ransac} for detection and removal of the support plane. 

\subsection{Evaluation Metrics}
Standard metrics for object segmentation are used, i.e., average precision, recall, and Intersection over Union (IoU). The association of predicted and ground-truth segments is performed by using the Hungarian algorithm~\cite{hungarian} where the cost of association is based on the IoU of two segments. To remove the factor of imperfect foreground detection, ground-truth foreground masks are provided for all methods, and only the segmentation of foreground pixels is evaluated. The segmentation results are tested on every 10 frames selected from the testing videos since neighboring frames in a 30 FPS video are very similar. The testing frames are selected from videos that are different from the training ones.

\subsection{Baseline Methods}
A set of baseline methods are considered for comparison. {\bf A.} DeepLabV3+~\cite{deeplabv3+} is a widely-used segmentation model for RGB images. This baseline is trained with supervision, which is considered an upper bound. The proposed method also uses DeepLabV3+ as a backbone network $\phi$ for pixel-wise feature extraction by removing its classification layer and adding a 16-dimensional embedding layer. {\bf B.} UOIS~\cite{uois} is an RGB-D object segmentation method designed for unseen tabletop objects. RICE~\cite{rice} is a follow-up work over UOIS, which refines its segmentation results. These methods are trained over a large amount of synthetic data with ground-truth annotations. Since they can barely be finetuned on datasets without ground-truth annotations, weights provided by the authors are directly used for testing. {\bf C.} LCCP~\cite{lccp} does object segmentation based on local convexity. Unlike its usage in the proposed method, which does segmentation on the whole reconstructed point cloud, here it performs segmentation on the point cloud that is converted from each testing frame using the camera intrinsic matrix. {\bf D.} SD-MaskRCNN~\cite{sd-maskrcnn} is a method for unknown object segmentation similar to UOIS. It is also trained with simulated data with ground truth but only takes depth images as input. {\bf E.} PiCIE~\cite{Cho_2021_CVPR} is an unsupervised semantic segmentation method using Invariance and Equivariance in Clustering. {\bf F.} STEGO~\cite{stego} is another unsupervised segmentation method, which distills correspondences between images into a set of class labels using a contrastive loss. The proposed method, PiCIE ({\bf E}) and STEGO ({\bf F}) are trained from scratch and tested using 3-fold cross-validation on the unlabeled videos, where background pixels are masked out.

\vspace{-.05in}
\begin{table}[h]
\centering
\begin{tabular}{|c|c|c|c|c|c|c|} 
\hline
     & \multicolumn{3}{c|}{Real Dataset} & \multicolumn{3}{c|}{Synthetic Dataset} \\
\hline
\hline
     {\bf Method} & {\bf Prec.} & {\bf Recall} & {\bf IoU} & {\bf Prec.} & {\bf Recall} & {\bf IoU}\\ 
\hline
   {\bf A.} \cite{deeplabv3+} & 0.948 & 0.947 & 0.910 & 0.977 & 0.932 & 0.915 \\
\hline
\hline
    {\bf B.} \cite{uois}+\cite{rice} & 0.859 & 0.753 & 0.711 & 0.963 & 0.854 & 0.838 \\
\hline
    {\bf C.} \cite{lccp} & 0.882 & 0.749 & 0.706 & {\bf 0.993} & 0.815 & 0.820\\
\hline
    {\bf D.} \cite{sd-maskrcnn} & 0.782 & 0.773 & 0.662 & 0.812 & 0.751 & 0.663\\
\hline
    {\bf E.} \cite{Cho_2021_CVPR} & 0.427 & 0.342 & 0.230 & 0.361 & 0.326 & 0.158\\
\hline
    {\bf F.} \cite{stego} & 0.318 & 0.343 & 0.162 & 0.325 & 0.337 & 0.142 \\
\hline
    \bf Ours & {\bf 0.925} & {\bf 0.911} & {\bf 0.870} & 0.938 & {\bf 0.925} & {\bf 0.880}\\
\hline
\end{tabular}
\caption{Segmentation results for different methods. Method A is a supervised solution provided as an upper bound of efficiency. The best results among unsupervised methods are in bold.}
\label{table:2}
\vspace{-.2in}
\end{table}

\subsection{Quantitative Results}
Quantitative results of the proposed and baseline methods are shown in Table~\ref{table:2}. The proposed method works well on both real and synthetic datasets. It outperforms all baseline methods in terms of mean IoU except {\bf A}, which was trained with supervision. {\bf B} and {\bf D} work well given that they are designed for unknown object segmentation and are only trained in simulation, but they can barely be improved without manual annotation on the test dataset. It is surprising that the unsupervised semantic segmentation methods {\bf E} and {\bf F} performed poorly on this task. This may be due to the lack of careful tuning of hyperparameters (default is used) and the lack of inductive bias provided for segmenting object instances. The work argues that learning based on geometric inductive bias and object reappearance across different scenes is effective for segmenting previously seen objects when deploying a robot in real environments.

\subsection{Intermediate 3D Segmentation Results}
\begin{figure}[t]%
    \centering
    {\includegraphics[width=\columnwidth]{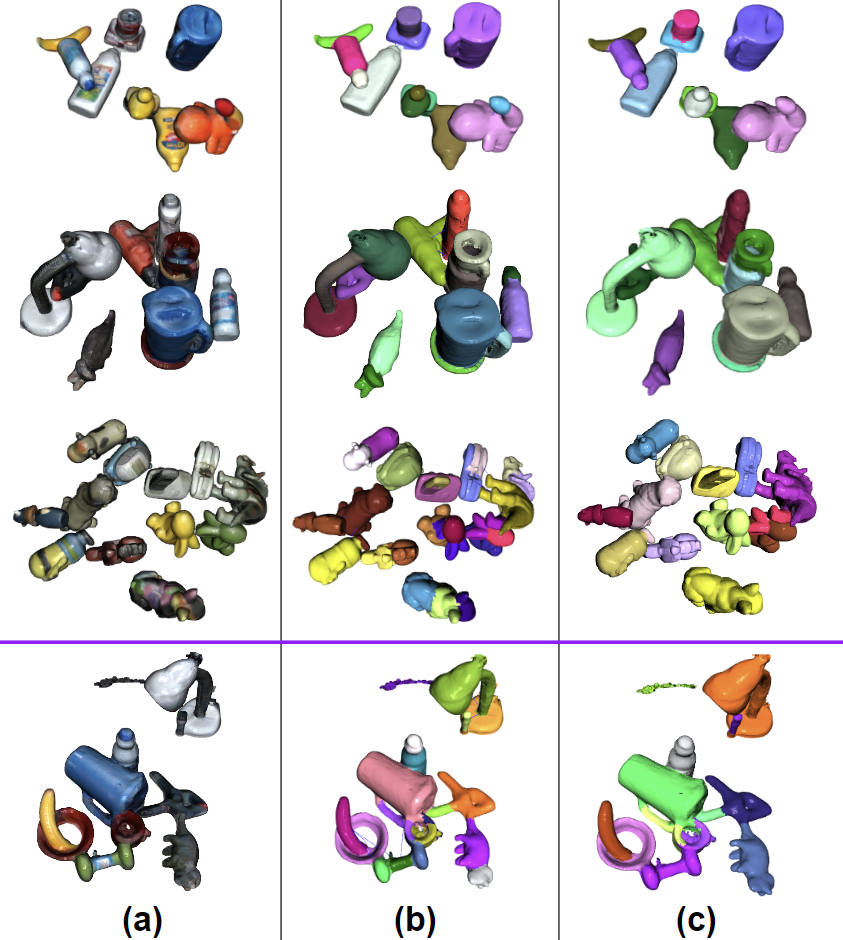}}
    \caption{\textbf{Intermediate Qualitative 3D Segmentation Results} \textbf{(a)} Reconstructed object point clouds in their original color. \textbf{(b)} Segmentation results from LCCP + SymSeg. \textbf{(c)} Object pseudo labels after graph matching and pruning given over-segmentation results as inputs. A failure case of over-segmentation is shown in the last row, where part of the "pitcher" and "cup" are not separated and the "cup" is segmented into inconsistent pieces. This cannot be addressed during graph matching.}%
    \label{fig:scene_level_results}%
    \vspace{-.25in}
\end{figure}

To assess the advantage of using graph matching for object pseudo-label generation, intermediate point cloud segmentation results are provided. The proposed method is compared against LCCP~\cite{lccp} and SymSeg~\cite{symseg} on the reconstructed object point cloud of each video. Intermediate quantitative and qualitative point cloud segmentation results are provided in Table~\ref{table:1} and Fig.~\ref{fig:scene_level_results} respectively. The benefits of combining LCCP with SymSeg are not entirely shown in numbers, as SymSeg sometimes can introduce object-level segmentation errors, such as over-segmenting a round cup and its handle. Nevertheless, it makes the 3D over-segmentation more consistent for cases like in Fig.~\ref{fig:split_and_merge}. The over-segmented objects, however, can be recovered by the graph-matching algorithm if an object pattern is found across scenes. The improvement brought by graph matching is even more pronounced over the more challenging geometries present in the synthetic dataset.

\begin{table}[h]
    \vspace{-.1in}
    \centering
    \begin{tabular}{|c|c|c|c|c|c|c|} 
    \hline
         & \multicolumn{3}{c|}{Real Dataset} & \multicolumn{3}{c|}{Synthetic Dataset} \\
    \hline
         & {\bf Prec.} & {\bf Recall} & {\bf IoU} & {\bf Prec.} & {\bf Recall} & {\bf IoU}\\ 
    \hline
        \cite{lccp} & {\bf 0.969} & 0.858 & 0.831 & 0.984 & 0.771 & 0.761\\
    \hline
        \cite{lccp} \& \cite{symseg} & 0.967 & 0.861 & 0.833 & {\bf 0.987} & 0.768 & 0.758\\
    \hline
        Ours & 0.957 & {\bf 0.964} & {\bf 0.923} & 0.978 & {\bf 0.966} & {\bf 0.950}\\
    \hline
    \end{tabular}
    \caption{Point Cloud Segmentation Results. "Ours" here corresponds to the object segmentation results achieved after graph matching given the input of over-segmentation from \cite{lccp} $\&$ \cite{symseg}. Best results in bold.} 
    \label{table:1}
    \vspace{-.15in}
\end{table}

\subsection{Ablation Study}
The results of an ablation study are shown in Table~\ref{table:3}. All the alternatives use the same feature extractor (DeepLabV3+~\cite{deeplabv3+}) and clustering method (MeanShift~\cite{banerjee2005clustering}). The ablations use different object pseudo-labels for training and are tested given ground truth foreground masks similar to Table~\ref{table:2}. {\bf V1.} The first one is trained on the COCO dataset~\cite{coco}. Its final prediction layer is removed and 256-dim pixel features are used directly for clustering. It has the lowest performance. {\bf V2.} The second one is trained using the output of LCCP~\cite{lccp} without graph matching. Its performance is similar to that of LCCP-based 3D segmentation as expected. {\bf V3.} The third one is trained using pseudo-labels with graph matching but without pruning. The performance is better than not using graph matching, but it's not as good as the proposed approach since nodes can be incorrectly merged, which results in erroneous object pseudo-labels. {\bf V4.} The last alternative tries to exhaustively register point clouds of connected components without using the candidates from graph matching. If the precision, recall, and feature cosine similarity of two registered nodes are greater than 0.9, then these two nodes are considered a match. This underperforms the proposed method since point cloud registration is more likely to fail due to clutter, and takes significantly more time to find object patterns.

\begin{table}[h]
\vspace{-.1in}
\centering
\begin{tabular}{|c|c|c|c|c|c|c|} 
\hline
     & \multicolumn{3}{c|}{Real Dataset} & \multicolumn{3}{c|}{Synthetic Dataset} \\
\hline
     {\bf Ablation} & {\bf Prec.} & {\bf Recall} & {\bf IoU} & {\bf Prec.} & {\bf Recall} & {\bf IoU}\\ 
\hline
    {\bf V1.} & 0.716 & 0.665 & 0.547 & 0.806 & 0.775 & 0.665\\
\hline
    {\bf V2.} & 0.915 & 0.815 & 0.776 & 0.935 & 0.746 & 0.711\\
\hline
    {\bf V3.} & 0.904 & 0.832 & 0.790 & 0.934 & 0.800 & 0.762 \\
\hline
    {\bf V4.} & 0.915 & 0.883 & 0.842 & 0.931 & 0.823 & 0.782 \\
\hline
    Ours & {\bf 0.925} & {\bf 0.911} & {\bf 0.870} & {\bf 0.938} & {\bf 0.925} & {\bf 0.880}\\
\hline
\end{tabular}
\vspace{-.05in}
\caption{Ablation results. Best results in bold.}
\label{table:3}
\vspace{-.2in}
\end{table}


\subsection{Running Time}
Binary linear programming (BLP) is an NP-hard problem. In practice, however, solutions can be acquired very fast given advanced solvers, such as Gurobi~\cite{gurobi}, which is used in companion implementation. The processes of graph matching (Sec.\ref{graph_matching}) and pruning (Sec.\ref{False Matches Pruning}) between two scenes can be done in about 2s and pairwise matching across 30 scenes can be done within $10$ minutes on an AMD Ryzen 5900 CPU.

\section{Conclusion and Limitations} \label{sec:limitation}
This work proposes a self-supervised learning pipeline given unlabeled RGB-D videos captured with a moving camera observing static scenes, which is a common scenario for household mobile robots. A graph-matching algorithm is adapted to find object patterns across videos and generate object pseudo-labels for learning. The proposed method achieves better results in object segmentation than existing unsupervised segmentation methods and methods that used simulation data for supervised training.

One limitation of the proposed approach is that it relies on the initial object over-segmentation to be consistent across different scenes. Despite using the synergy of LCCP and SymSeg, consistent over-segmentation may not always be achieved when reconstruction noise is large and when objects are not able to be separated using geometric features. This can potentially be addressed with more data that cover multiple over-segmentation patterns and future improvements on the matching algorithm that considers the operation of splitting a node when under-segmentation happens. Another limitation is that the proposed method cannot handle deformable objects in general, since it requires matched objects to be identical in both texture and geometry. 

\pagebreak
\bibliographystyle{IEEEtran}
\bibliography{reference}


\end{document}